\documentclass[10pt,twocolumn,letterpaper]{article}

\usepackage[applications]{wacv}              % CAMERA-READY version, Applications track
\usepackage{multirow}
\usepackage{adjustbox}
\usepackage[accsupp]{axessibility}
\usepackage{microtype}
\usepackage{makecell}
\newcolumntype{A}[1]{>{\raggedright\arraybackslash}p{#1}}

\makeatletter
\renewcommand\paragraph{%
  \@startsection{paragraph}{4}{\z@}%
    {0.8ex \@plus 0.2ex \@minus 0.1ex}%
    {0.4em}%
    {\normalfont\normalsize\bfseries}}
\makeatother

\AtBeginEnvironment{table}{%
  \footnotesize
  \setlength{\tabcolsep}{2pt}%
}

\definecolor{wacvblue}{rgb}{0.21,0.49,0.74}
\usepackage[pagebackref,breaklinks,colorlinks,allcolors=wacvblue]{hyperref}

\def\wacvPaperID{1294}
\def\confName{WACV}
\def\confYear{2027}

\begin{document}

\title{AnimeAdapter: A Modular Adapter for Appearance-Consistent Anime Character Generation}

\author{Yixuan Han\\
National University of Singapore, Singapore\\
{\tt\small e1143645@u.nus.edu}}

\maketitle

\begin{abstract}
  \sloppy
We present AnimeAdapter, a lightweight appearance adapter for Stable Diffusion that enables controllable, appearance-consistent anime character editing from a single reference image. The system injects fine-grained CLIP patch tokens into U-Net decoupled cross-attention, with token-level foreground masking and pose-guided training to disentangle appearance from layout. Once pretrained on curated Danbooru-style data, the adapter requires no per-subject fine-tuning at deployment and remains compatible with ControlNet, T2I-Adapter, and LoRA. We further introduce a structured anime character editing benchmark with taxonomy-guided prompts, OpenPose conditions, and token-level masks, covering pose, expression, scene, and local motion editing. Experiments on our benchmark and DreamBench++ show improved appearance preservation over IP-Adapter variants and several subject-driven baselines, with qualitative demonstrations for comic-style production workflows. Code and weights will be released.
\end{abstract}

\begin{figure*}[t]
  \centering
  \includegraphics[width=12.2cm]{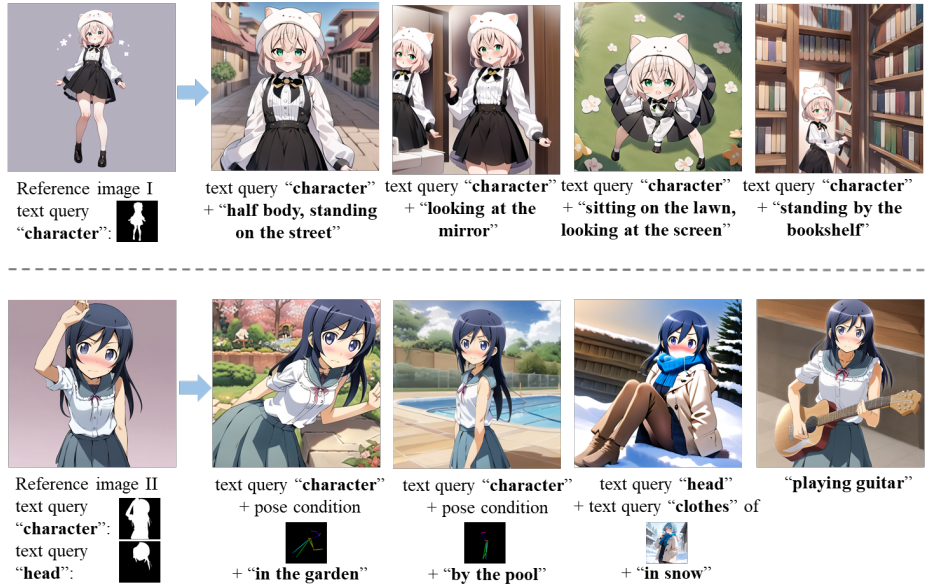}
  \caption{Results of our method. AnimeAdapter is a lightweight adapter designed to enable appearance-consistent generation of anime characters. Without additional per-subject training at deployment time, it supports arbitrary anime subject-driven generation in a zero-shot manner (no test-time fine-tuning on the reference) and remains fully compatible with the Stable Diffusion ecosystem. Its unique architecture introduces semantic local attention with token-level masking, enabling selective preservation of appearance details from reference images. This capability facilitates automated offline comic book production.}
  \label{fig:first}
\end{figure*}

\section{Introduction}

Anime and comic production relies heavily on redrawing the same character across many panels, where pose, expression, scene, and local actions change while the character's identity must stay visually consistent. In practice this means preserving fine-grained appearance attributes, hair highlights, outfit patterns, color schemes, and body shape, across diverse edits. Modern text-to-image (T2I) diffusion models such as Stable Diffusion (SD)~\cite{rombach2022high} produce high-quality illustrations from text, and reference-based adapters such as IP-Adapter~\cite{IP-Adapter} further align generation with a reference image. However, text alone cannot specify such fine-grained appearance, and generic adapters tend to capture global semantics while blurring the subtle, identity-defining details that matter most for anime characters. For an automated comic workflow, what creators need is a plug-and-play module that preserves these attributes from a single reference, supports common controls such as ControlNet and LoRA, and avoids costly per-character fine-tuning at deployment.

Existing solutions fall short of this goal in different ways. Per-subject personalization methods~\cite{DreamBooth,textualinversion,hu2022lora} achieve strong identity fidelity but require optimizing new weights or tokens for every character, which is impractical for large-scale production. Fine-tuning-free encoders~\cite{IP-Adapter,Ssr-encoder,InstantID} instead inject reference features through decoupled cross-attention in a single forward pass, but they are designed for general objects or human faces and provide no mechanism to selectively preserve localized anime appearance. Moreover, when appearance conditioning is strengthened, these methods often entangle appearance with the reference layout, causing the model to copy the reference pose rather than follow the editing prompt.

We address these issues with AnimeAdapter, a lightweight adapter for SD that targets appearance-consistent anime character editing as a deployable component. AnimeAdapter injects fine-grained CLIP patch tokens into the U-Net through decoupled cross-attention, and introduces semantic-selective local attention based on token-level foreground masks, so that appearance is preserved only from the subject region. To prevent appearance–layout entanglement, we adopt a pose-guided training strategy (\cref{fig:disen}) that offloads structural guidance to a frozen pose controller, letting the adapter focus on appearance. Trained once on curated Danbooru-style data, the module generalizes to unseen characters from a single reference without per-subject fine-tuning, and remains compatible with ControlNet, T2I-Adapter, and LoRA. We further build a structured benchmark for anime character editing, enabling systematic evaluation across pose, expression, scene, and local-motion edits. Representative results are shown in \cref{fig:first} and \cref{fig:application}.

Our contributions are summarized as follows:
\begin{itemize}
\item We present AnimeAdapter, a pluggable SD adapter for appearance-consistent anime character editing that works from a single reference image, requires no per-subject fine-tuning at deployment, and integrates with ControlNet, T2I-Adapter, and LoRA.
\item We introduce token-level masked appearance injection together with a pose-disentangled training strategy, which selectively preserves fine-grained character attributes while enabling smooth and controllable local and global editing.
\item We construct a high-resolution anime character benchmark from SFW Danbooru-style tags, with a taxonomy-guided prompt design, OpenPose conditions, and token-level masks, supporting reproducible evaluation across diverse editing scenarios.
\end{itemize}

\begin{figure*}[t]
  \centering
  \begin{subfigure}{0.49\textwidth}
    \centering
    \includegraphics[width=\linewidth]{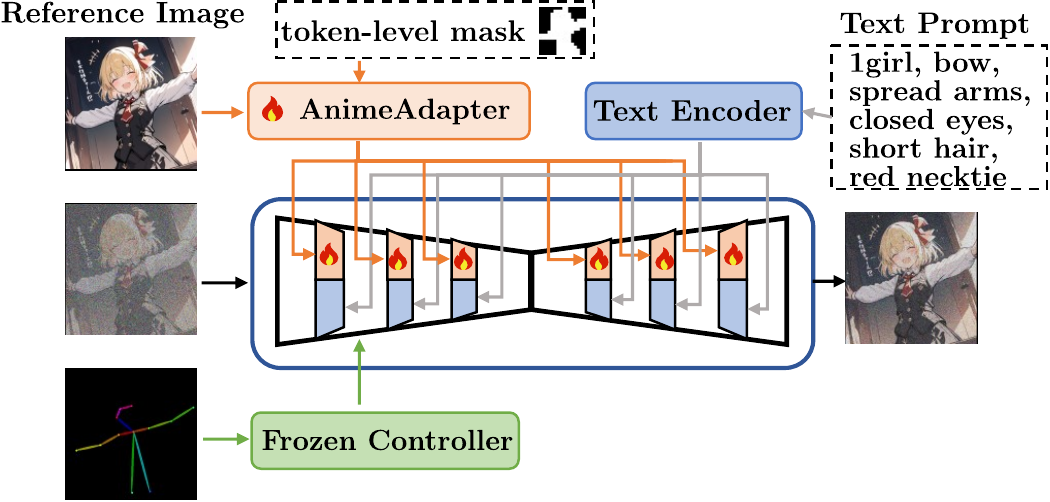}
    \caption{Training sample with prompt, mask, and OpenPose.}
    \label{fig:main_train}
  \end{subfigure}
  \hfill
  \begin{subfigure}{0.49\textwidth}
    \centering
    \includegraphics[width=\linewidth]{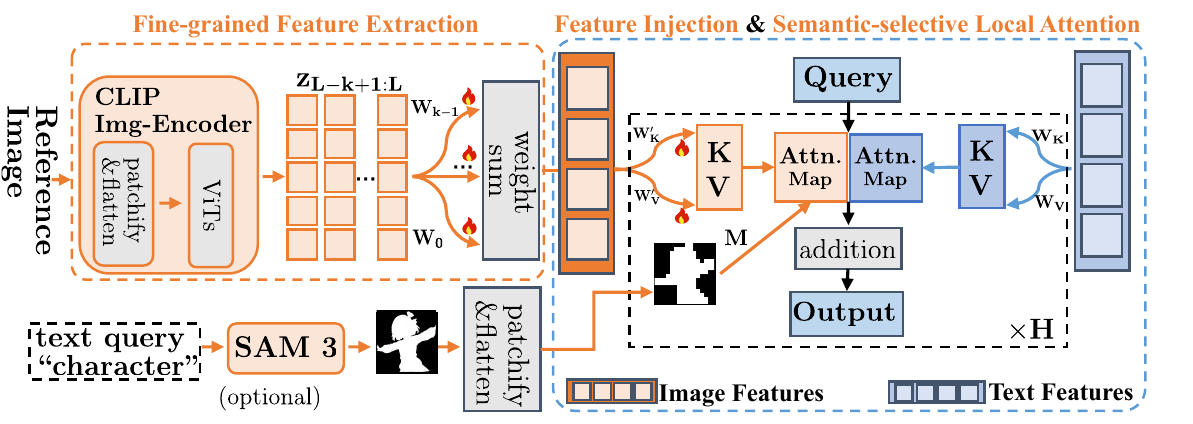}
    \caption{Fine-grained feature extraction and injection.}
    \label{fig:main_inject}
  \end{subfigure}
  \caption{Overview of the proposed framework. The left panel shows that each training sample contains an image, text prompt, subject token-level mask, and OpenPose condition. The reference image is processed into fine-grained tokens via a CLIP image encoder, which are injected into the U-Net through decoupled cross-attention. The right panel illustrates the details of fine-grained feature extraction and injection, in which the token-level masks are obtained using text-driven segmentation from SAM~3.}
  \label{fig:main}
\end{figure*}

\section{Related Work}

\subsection{Subject-Driven Generation and Image Conditioning}
Built on text-to-image diffusion models~\cite{ho2020denoising,nichol2021glide,saharia2022photorealistic,ramesh2021zero,ramesh2022hierarchical,peebles2023scalable}, in particular Stable Diffusion~\cite{rombach2022high}, and on structural controllers~\cite{zhang2023adding,T2I-Adapter,li2023gligen,xie2023boxdiff}, recent work targets fine-grained visual control that text prompts alone cannot specify, motivating reference-based and structurally conditioned generation. One line of work personalizes a pretrained model to a specific subject. DreamBooth~\cite{DreamBooth} and Textual Inversion~\cite{textualinversion} bind a subject to rare tokens or a pseudo-word, while LoRA~\cite{hu2022lora} injects concept-specific information through trainable low-rank matrices; follow-up methods~\cite{arar2023domain,gu2023mix,hao2023vico,ruiz2024hyperdreambooth,smith2023continual} improve efficiency. These approaches achieve strong identity fidelity but require per-subject optimization, which is costly for large-scale comic production.

To avoid per-subject fine-tuning, fine-tuning-free encoders inject reference features in a single forward pass. IP-Adapter~\cite{IP-Adapter} introduces a decoupled cross-attention layer that adds image-conditioned attention on top of the original text cross-attention, and SSR-Encoder~\cite{Ssr-encoder}, InstantID~\cite{InstantID}, and Face-Adapter~\cite{han2024face} follow this paradigm for selective subjects or human faces. A complementary family modifies self-attention to reuse reference features: MasaCtrl~\cite{MasaCtrl} queries an inverted source pipeline, and DreamTuner~\cite{Dreamtuner} and related methods~\cite{Magicanimate,Animateanyone} concatenate reference features into self-attention keys and values, typically requiring a trainable or frozen U-Net copy. Structural conditioning is often combined with these adapters: ControlNet~\cite{zhang2023adding} and T2I-Adapter~\cite{T2I-Adapter} inject pose, edge, or depth signals to control layout. Our adapter builds on the lightweight decoupled cross-attention design, but unlike these general-purpose methods it targets fine-grained anime appearance through token-level masked injection, and it can be combined with ControlNet and T2I-Adapter as frozen controllers.

\subsection{Anime Character Consistency and Editing}
Maintaining character appearance across contexts has attracted growing attention. Many appearance-consistent methods~\cite{Animateanyone,Magicanimate,Disco,shi2024dragdiffusion,zhou2024storydiffusion,chen2024anydoor,shen2023advancing} rely on video or illustration datasets that provide multiple views of the same character under varying poses and backgrounds, which helps disentangle appearance from context but assumes multi-frame supervision rather than single-reference editing. Other works~\cite{ai2023stable} mine similar anime image pairs as pseudo ground truth to preserve high-fidelity appearance, and NeRF-based approaches~\cite{patashnik2024consolidating} use 2D-to-3D priors to enhance character-specific details. In contrast, we derive each training pair from a single image and rely on the generative capacity of SD to integrate detailed appearance, focusing specifically on plug-and-play anime character editing. Compared with prior work, existing methods leave three gaps for anime production: they rarely preserve fine-grained localized appearance, they lack a mechanism for selective region-level conditioning, and they offer no structured benchmark for taxonomy-guided anime character editing, all of which we address in this work.

\begin{figure*}[t]
  \centering
  \includegraphics[width=12.2cm]{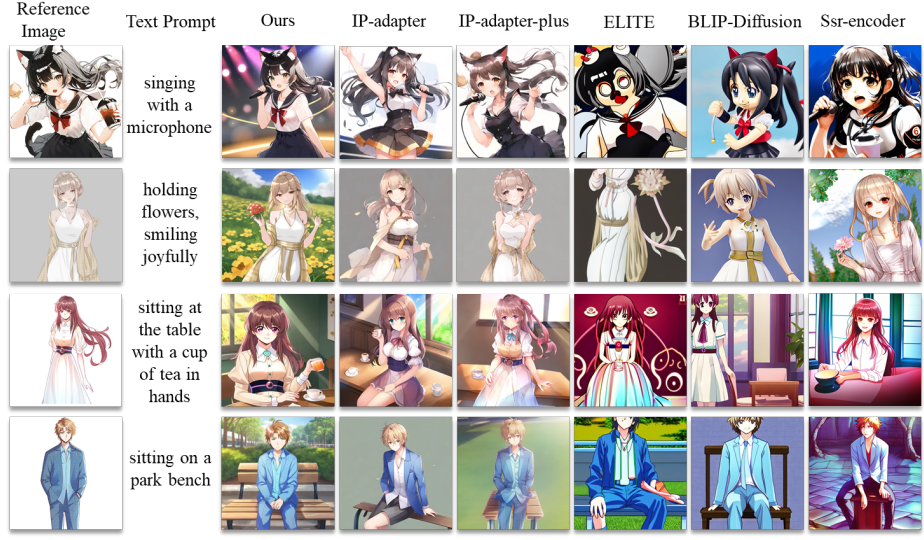}
  \caption{Qualitative comparison of different methods. Our framework has higher performance in terms of anime character appearance consistency.}
  \label{fig:comparison}
\end{figure*}

\section{Method}

\subsection{Fine-grained Feature Extraction}
We adopt CLIP as our image encoder because its features are semantically aligned with the Stable Diffusion text encoder, which facilitates injection into the cross-attention space.

Given a preprocessed reference RGB image $\mathbf{x} \in \mathbb{R}^{H \times W \times 3}$, the image is partitioned into non-overlapping $p \times p$ patches.
Each patch is mapped into a learnable embedding space, and a class embedding $\boldsymbol{[\mathrm{CLS}]}$ is prepended to capture global semantics.
After adding positional encodings, the resulting token sequence
$
\mathbf{z}_0 \in \mathbb{R}^{N \times D}
$
is fed into a stack of $L$ Vision Transformer (ViT) layers, where $N = 1 + \frac{HW}{p^2}$.
The output of each layer is defined as
\begin{equation}
\mathbf{z}_\ell = \mathrm{ViT}_\ell(\mathbf{z}_{\ell-1}), \quad \ell = 1, 2, \dots, L,
\end{equation}
where $\mathbf{z}_\ell \in \mathbb{R}^{N \times D}$ and $D$ denotes the hidden dimension of the CLIP image encoder.

Following prior works~\cite{Ssr-encoder, gandelsman2023interpreting}, we aggregate features from the last $k$ ViT layers rather than from a single block: deeper CLIP layers progressively encode spatially meaningful patch--token correspondence, which is important for localized appearance control. Each selected layer is linearly projected into the Stable Diffusion cross-attention space and fused via a learnable layer-scale mechanism:
\begin{equation}
\label{eq:clip-fuse}
\mathbf{I} = \mathrm{LN}\!\left(\sum_{i=0}^{k-1} \alpha_i \cdot \mathbf{z}_{L-i} \mathbf{W}_i \right),
\end{equation}
where $\alpha_i$ are learnable scalars, $\mathbf{W}_i \in \mathbb{R}^{D \times D'}$ are linear projections, and $D'$ matches the cross-attention dimension of the U-Net.
All operations are token-wise and do not introduce cross-token coupling, thereby preserving the spatial alignment needed for token-level masking (\cref{sec:local-attn}).
We use $k{=}7$ in all main experiments; an ablation over layer depth is reported in \cref{tab:ablation_clip}.

\subsection{Fine-grained Feature Injection}
As shown in \cref{fig:main}, we inject the fine-grained image features $\mathbf{I} \in \mathbb{R}^{N \times D'}$ into the U-Net using a decoupled cross-attention mechanism, following IP-Adapter~\cite{IP-Adapter}.
For each cross-attention layer, the output is computed as
\begin{equation}
\mathbf{Z} = \mathrm{Attn}(\mathbf{Q}, \mathbf{K}, \mathbf{V}) 
+ \gamma \cdot \mathrm{Attn}(\mathbf{Q}, \mathbf{I}\mathbf{W}_K', \mathbf{I}\mathbf{W}_V'),
\end{equation}
where $\mathbf{Q}$, $\mathbf{K}$, and $\mathbf{V}$ denote the original query, key, and value tokens, respectively, and $\mathbf{W}_K', \mathbf{W}_V'$ are learnable projections for image features.
This is the single-reference injection form; \cref{sec:local-attn} extends it with token-level masks and, optionally, multiple references at inference.

\subsection{Semantic-selective Local Attention}
\label{sec:local-attn}

The spatially aligned CLIP tokens above enable semantic-selective local attention: instead of injecting appearance from the entire reference, we restrict the image branch to subject regions indicated by token-level masks, so that background pixels do not leak into the generation.

\begin{figure*}[t]
  \centering
  \includegraphics[width=12.2cm]{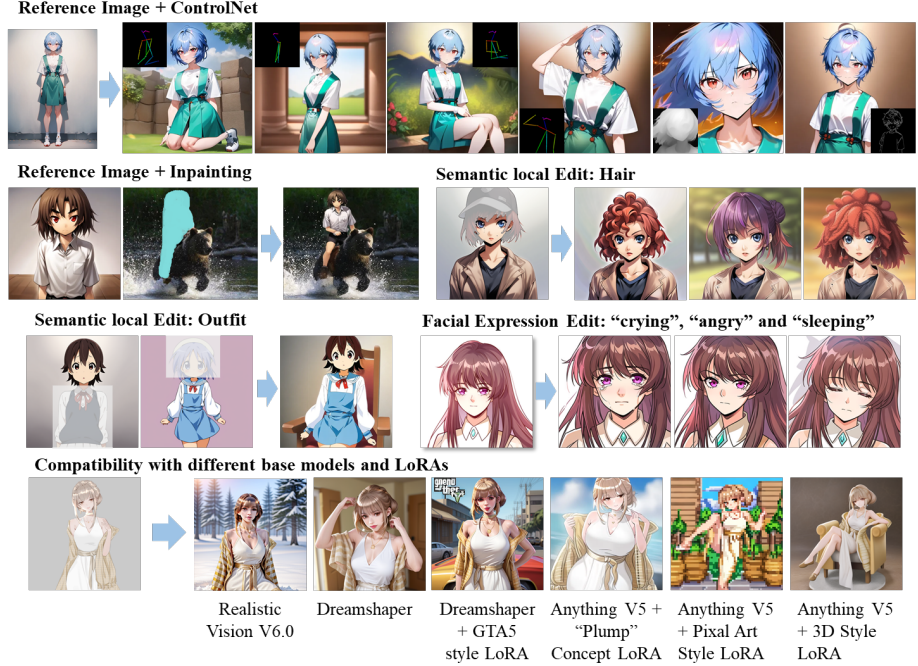}
  \caption{Qualitative results demonstrating the versatility of AnimeAdapter. Our method enables smooth integration with additional concept conditions and preserves appearance selectively from different reference images. It also shows compatibility with different base models and LoRAs.}
  \label{fig:application}
\end{figure*}

\paragraph{Training and inference settings.}
During training, each sample uses the same image as both the reconstruction target and the reference; we extract a single feature map $\mathbf{I}$ as in \cref{eq:clip-fuse} and apply the corresponding foreground mask.
At inference, our default setting is also \textbf{single-reference} appearance-consistent editing: one reference image yields $\mathbf{I}_1=\mathbf{I}$.
As an optional extension, multiple reference images can be supplied at inference; each reference $i$ is encoded into $\mathbf{I}_i$ with its own mask, and appearance is injected only from the unmasked tokens of each reference.
This supports semantic local editing, where different subject regions or entities are borrowed from different references (see the semantic local editing examples in \cref{fig:application}).

\paragraph{Masked image-branch attention.}
Let $\mathbf{I}_i \in \mathbb{R}^{N \times D'}$ denote the fine-grained tokens extracted from the $i$-th reference image via \cref{eq:clip-fuse} (so $\mathbf{I}_1=\mathbf{I}$ when only one reference is used).
We associate each reference with a token-level mask that marks the subject region: background tokens are suppressed and foreground tokens carry the appearance to preserve.
The masked injection is
\begin{equation}
\mathbf{Z} = \mathrm{Attn}(\mathbf{Q}, \mathbf{K}, \mathbf{V}) + \sum_i \gamma_i' \cdot \mathbf{O}_i,
\end{equation}
where $\gamma_i'$ is a guidance scale for the $i$-th reference and $\mathbf{O}_i$ is the masked image-branch output defined below.
Rather than masking the features directly, we apply the mask on the image-branch attention scores $\mathbf{Q}(\mathbf{I}_i \mathbf{W}_K')^\top$, and adopt different strategies during training and inference. Let $\mathbf{L}_i = \mathbf{Q}(\mathbf{I}_i \mathbf{W}_K')^\top / \sqrt{d'}$ be the image-branch logits of a cross-attention layer, where $d'$ is the head dimension. During training, we add a large negative bias $\mathbf{B}_i$ on the masked (background) keys before the softmax, so that these regions cannot contribute to the aggregated appearance:
\begin{equation}
\mathbf{P}_i^{\mathrm{tr}} = \mathrm{softmax}\!\left(\mathbf{L}_i + \mathbf{B}_i\right), \qquad
\mathbf{O}_i^{\mathrm{tr}} = \mathbf{P}_i^{\mathrm{tr}}\,(\mathbf{I}_i \mathbf{W}_V').
\end{equation}
\vspace{-0.35em}\noindent
During inference, applying the same logit bias tends to lower image quality and introduce artifacts; we therefore keep the full softmax and instead apply a binary mask $\mathbf{T}_i$ multiplicatively after the softmax:
\begin{equation}
\mathbf{P}_i^{\mathrm{inf}} = \mathrm{softmax}\!\left(\mathbf{L}_i\right), \qquad
\mathbf{O}_i^{\mathrm{inf}} = \left(\mathbf{P}_i^{\mathrm{inf}} \odot \mathbf{T}_i\right)(\mathbf{I}_i \mathbf{W}_V').
\end{equation}
In both cases, suppressing the background keys lets the model preserve appearance only from the subject region, enabling localized appearance control. Quantitative results in \cref{tab:ablation_mask} show that this mechanism yields stronger subject appearance preservation.

We employ SAM~3~\cite{SAM3} to segment subject regions from reference images based on prompts, producing the token-level masks used in the attention mechanism above (\cref{fig:main_inject}).

\subsection{Strategy of Pose/Layout Disentanglement}

\begin{figure*}[t]
  \centering
  \includegraphics[width=12.2cm]{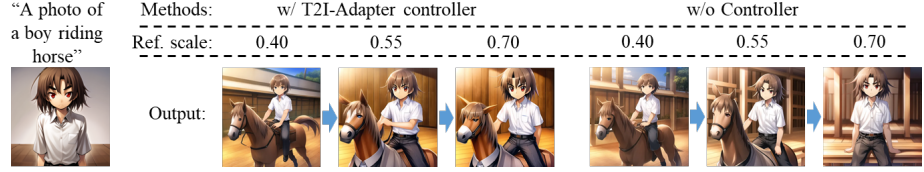}
  \caption{Qualitative results demonstrating our pose/layout disentanglement training strategy.}
  \label{fig:disen}
\end{figure*}

A common issue in our initially trained model is the entanglement between appearance and layout information. As shown on the right side of \cref{fig:disen}, increasing the strength of appearance conditioning causes the model to unintentionally replicate the reference layout, rather than following the editing prompt. Such layout leakage undermines controllability and limits structural flexibility.

To address this issue, we explicitly introduce a frozen pose controller during training to disentangle appearance and layout information. 
The core idea is to offload structural guidance to an external module, so that the adapter focuses on extracting appearance-specific representations.

Concretely, during each training step, we extract pose conditions from the reference image and feed them into a frozen structural controller. 
The target task remains image reconstruction; however, the layout information is now provided by the controller rather than being implicitly learned by the adapter. 
This design encourages the adapter to avoid redundantly encoding pose information and instead concentrate on learning effective appearance features.

In our experiments, we adopt the OpenPose-conditioned T2I-Adapter and ControlNet as controllers. As shown in \cref{tab:ablation_controller}, a frozen T2I-Adapter enables AnimeAdapter to better preserve appearance while achieving stronger pose editing capability. The left side of \cref{fig:disen} shows the effect of our training strategy.

\section{Dataset Construction}
\subsection{Motivation and Overview}
Our training objective is formulated as an image reconstruction task, where the model reconstructs a reference image while injecting fine-grained character appearance information via the proposed adapter (\cref{fig:main_train}).
To support both training and evaluation, we construct a high-quality anime character dataset consisting of reference images, structured prompts, OpenPose conditions, and token-level character masks. The dataset is specifically designed to enable controlled appearance injection and systematic evaluation of text-driven editing capabilities.
Full taxonomy definitions, reference/editing prompt templates, and per-task construction rules are provided in the supplement (Detailed Evaluation Dataset Construction).
\subsection{Data Source and Preprocessing}
Our dataset is built upon Danbooru2025 metadata~\cite{danbooru2025metadata}, which contains approximately 9.1M entries annotated with rich semantic tags. Each entry provides four tag categories: general descriptive tags, character identity tags, artist tags, and a content rating label.
After filtering for safe-for-work (SFW) content and single-character entries, about 4.5M metadata samples remain for training and evaluation.

\subsection{Semantic Tag Reorganization}
To enable evaluation of structured semantic editing, we reorganize approximately 4,600 frequently occurring Danbooru tags into a structured semantic taxonomy.
Specifically:

\begin{itemize}
\item \textbf{Cluster 0: Identity and count.} Character identity indicators and character count tags (e.g., \texttt{solo}, \texttt{1girl}).
\item \textbf{Cluster 1: View and framing.} Camera viewpoint, perspective, and body framing attributes.
\item \textbf{Cluster 2: Style and aesthetics.} Rendering styles, lighting, background, and artistic techniques.
\item \textbf{Cluster 3: Global posture.} Whole-body poses and motion states.
\item \textbf{Cluster 4: Local body-part attributes.} Fine-grained descriptions. This cluster is further organized hierarchically according to anatomical regions and wearable components. Within each sub-group, tags are divided into attribute-type tags, which describe static visual properties (e.g., clothing type or accessory presence), and motion-type tags, which describe localized actions or dynamic states (e.g., hand gestures or facial movements). 
\item \textbf{Cluster 5: Visual patterns.} Pattern-related structural attributes.
\end{itemize}

This structured taxonomy enables semantic-level manipulation while preserving consistency across unrelated attributes.

\subsection{Training Prompt Construction}

For training, we construct two prompts per metadata entry: the Reference Image Generation Prompt, which includes clusters 0--5 plus the character name to capture full appearance details, and the Training Prompt, which uses clusters 1--5 only, omitting character name and rating.

\subsection{Evaluation Dataset}
We construct a structured evaluation dataset designed to measure appearance preservation, semantic editability, and pose controllability under systematically defined editing tasks.
Unlike conventional editing benchmarks where edited prompts are directly derived from the original prompt through simple modifications, our evaluation protocol constructs task-specific reference images and corresponding editing prompts independently from the same underlying semantic metadata.

\paragraph{Task-Specific Reference Image Construction.}
For each metadata entry, we first reorganize Danbooru tags into semantic clusters as described above. Based on these clusters, we construct four types of task-specific reference prompts, each designed to isolate a particular aspect of character representation: an original-appearance reference, a full-body reference, an upper-body reference, and a portrait reference. Each reference prompt selectively includes identity, framing, and appearance attributes to produce a reference image tailored to a specific evaluation objective.

Reference images for both training and evaluation are generated at resolution $1024 \times 1024$ using Animagine XL 4.0~\cite{animaginexl4}, a diffusion model trained on Danbooru-style anime data.

For each reference image, we additionally extract a token-level foreground mask using SAM~3~\cite{SAM3} and an OpenPose skeleton, both used for training and evaluation.

\paragraph{Evaluation Task Types.}
Each evaluation task consists of a reference image and a corresponding editing prompt, targeting a specific editing capability. The introduction of each evaluation task and its editing prompt design are as follows:

\textbf{Body-part Motion Editing.} Reference image: original appearance reference. Editing prompt: replaces body-part motion-related tags using taxonomy-aware substitution within Cluster 4.

\textbf{Text-driven Posture and Viewpoint Editing.} Reference image: full-body or upper-body reference. Editing prompt: modifies body-related structural attributes, including posture, viewpoint, perspective, framing, and depth-related tags.

\textbf{Facial Expression Editing.} Reference image: portrait reference. Editing prompt: introduces new facial expression attributes.

\textbf{Scene Editing.} Reference image: full-body reference. Editing prompt: introduces scene descriptions.

\textbf{Pose-conditioned Editing.} Reference image: full-body reference. Editing prompt: posture and framing modification combined with OpenPose condition.

\paragraph{Advantages of Structured Evaluation Construction.}
This protocol offers three benefits: (1) disentangled evaluation across semantic editing dimensions, (2) reduced prompt conflicts with consistent reference-edit pairing, and (3) fully reproducible and scalable benchmarking.

\section{Experiments}

\subsection{Metrics}

We train AnimeAdapter for 30{,}000 steps on 24{,}000 samples (effective batch size 16 on an A100); only ${\sim}$24.7M newly introduced adapter weights are updated while the SD backbone remains frozen.
The classifier-free guidance dropout schedule for text and image conditions is given in the supplement (Training CFG Setting).
To comprehensively evaluate appearance preservation, text alignment, diversity, and pose controllability, we adopt multiple quantitative metrics. 
Unless otherwise specified, we generate four samples per input and report averaged results.

{\bf Appearance Preservation.}
We compute CLIP image similarity between the generated image and the reference image. 
To remove background bias, we use SAM~3 to segment foreground characters and replace non-character regions with white before computing similarity. 
This masked CLIP-I metric follows~\cite{clip_i_cc_2024}. 
We further report LPIPS~\cite{zhang2018lpips} and PSNR on masked foreground regions.

{\bf Text Alignment and Diversity.}
We adopt CLIP-T~\cite{hessel2021clipscore} to evaluate text-image consistency. 
Following DreamBooth~\cite{DreamBooth}, we measure diversity by the pairwise LPIPS among four generated samples; ``Div.'' denotes this LPIPS-div metric, where higher values indicate more diverse generations.

{\bf Pose Control.}
For pose-conditioned experiments, ``AKD'' denotes Average Keypoint Distance and ``MKR'' denotes Missing Keypoint Rate between the detected and target skeletons, following MagicAnimate~\cite{Magicanimate}. We additionally report ``Fail'', which denotes the fraction of generated images on which OpenPose cannot reliably detect a pose, reflecting how often the generation degrades under pose conditioning.

\subsection{Comparison}

{\bf Quantitative Comparison.}
\cref{tab:quantitative} summarizes macro-averaged scores over our editing tasks; complete per-task breakdowns are reported in the supplement (Total Quantitative Results).
Each assessment task includes around 2000 test cases.
Our method consistently achieves higher CLIP-I and lower LPIPS, demonstrating improved appearance preservation. It also attains a lower failure rate while maintaining low AKD and MKR, indicating that stronger appearance conditioning does not come at the cost of pose controllability. ``Ours'' denotes our full model (Full+Mask+Ctrl).

\begin{table}[t]
\centering
\caption{Quantitative comparison with other models.}
\label{tab:quantitative}
\begin{subtable}{\columnwidth}
  \centering
  \caption{Appearance preservation.}
  \label{tab:quantitative_app}
  \begin{adjustbox}{max width=\columnwidth}
  \begin{tabular}{@{}lccccc@{}}
  \toprule
  Method & CLIP-T $\uparrow$ & CLIP-I $\uparrow$ & LPIPS $\downarrow$ & Div. $\uparrow$ & PSNR $\uparrow$ \\
  \midrule
  IP-Adapter & 0.179 & 0.791 & 0.503 & \textbf{0.473} & 10.173 \\
  IP-Adapter Plus & 0.175 & 0.815 & 0.434 & 0.378 & 11.288 \\
  Ours & \textbf{0.187} & \textbf{0.860} & \textbf{0.431} & 0.422 & \textbf{11.437} \\
  \bottomrule
  \end{tabular}
  \end{adjustbox}
\end{subtable}

\vspace{0.3em}
\begin{subtable}{\columnwidth}
  \centering
  \caption{Pose controllability.}
  \label{tab:quantitative_pose}
  \begin{adjustbox}{max width=\columnwidth}
  \begin{tabular}{@{}lccc@{}}
  \toprule
  Method & AKD $\downarrow$ & MKR $\downarrow$ & Fail $\downarrow$ \\
  \midrule
  IP-Adapter & 0.109 & 0.106 & 0.550 \\
  IP-Adapter Plus & \textbf{0.070} & 0.075 & 0.273 \\
  Ours & 0.090 & \textbf{0.071} & \textbf{0.255} \\
  \bottomrule
  \end{tabular}
  \end{adjustbox}
\end{subtable}
\end{table}

{\bf Qualitative Comparison.}
\cref{fig:comparison} presents visual comparisons against IP-Adapter variants. Our model better preserves fine-grained, identity-defining attributes such as hair color and highlights, clothing textures, and accessory details, while still following the editing prompt. In contrast, the baselines tend to retain only the global semantics of the reference and blur these local details.

{\bf Public Benchmark Evaluation.}
To evaluate generalization beyond our own benchmark, we further compare against a broader set of subject-driven baselines on DreamBench++~\cite{peng2025iclr-dreambench}, using its live-subject human split (nine cartoon-style identities) with shared prompts and the same metrics as above. As reported in \cref{tab:dreambench}, AnimeAdapter attains the best CLIP-T and CLIP-I together with the highest PSNR and competitive LPIPS-div, confirming that our fine-grained appearance injection transfers to an external, human-aligned benchmark rather than overfitting to our data distribution.

\begin{table}[t]
\centering
\caption{Quantitative comparison on the DreamBench++ human cartoon subset (nine identities).}
\label{tab:dreambench}
\begin{adjustbox}{max width=\columnwidth}
\begin{tabular}{@{}lccccc@{}}
\toprule
Method & CLIP-T $\uparrow$ & CLIP-I $\uparrow$ & LPIPS $\downarrow$ & Div. $\uparrow$ & PSNR $\uparrow$ \\
\midrule
IP-Adapter (Huge14) & 0.216 & 0.755 & 0.483 & 0.452 & 9.615 \\
IP-Adapter (BigG) & 0.206 & 0.748 & 0.478 & 0.470 & 9.724 \\
IP-Adapter Plus (Huge14) & 0.179 & 0.803 & \textbf{0.375} & 0.426 & 10.744 \\
SSR-Encoder & 0.211 & 0.798 & 0.496 & 0.510 & 8.764 \\
BLIP-Diffusion & 0.200 & 0.712 & 0.640 & \textbf{0.533} & 7.899 \\
ELITE & 0.222 & 0.754 & 0.533 & 0.530 & 8.117 \\
Ours & \textbf{0.237} & \textbf{0.851} & 0.390 & 0.505 & \textbf{11.272} \\
\bottomrule
\end{tabular}
\end{adjustbox}
\end{table}

\subsection{Ablation Study}

All ablation experiments are conducted using CLIP ViT-L/14 as the vision backbone. 
Each variant is trained on the same 5,000-image training set for an identical number of iterations. 
Evaluation is performed on our test set, where each assessment task includes 100 test cases.
In \cref{tab:ablation_injection,tab:ablation_mask,tab:ablation_controller}, each Combination label concatenates three factors: ``full'' denotes Full-Blocks injection and ``up'' denotes Up-Blocks-only injection; ``w/o disent.'' denotes no disentanglement controller, while ``T2I-Adapter'' and ``ControlNet'' denote the frozen pose controller used otherwise; ``w.\ mask'' and ``w/o mask'' denote with and without token-level masking.

{\bf Injection Scope.}
We compare Up-Blocks-only injection and Full-Blocks injection in \cref{tab:ablation_injection}. 
While InstantStyle~\cite{instantstyle} observes that reference style information leaks mainly into the upsampling blocks, we find that preserving fine-grained appearance benefits from broader injection: Full-Blocks injection consistently achieves higher CLIP-I and lower LPIPS across all controller settings.

\begin{table}[t]
\centering
\caption{Effect of feature injection scope.}
\label{tab:ablation_injection}
\begin{adjustbox}{max width=\columnwidth}
\begin{tabular}{@{}lcccc@{}}
\toprule
Combination & CLIP-T $\uparrow$ & CLIP-I $\uparrow$ & LPIPS $\downarrow$ & PSNR $\uparrow$ \\
\midrule
full+ w/o disent.+ w. mask & 0.188 & \textbf{0.829} & \textbf{0.493} & \textbf{9.697} \\
up+ w/o disent.+ w. mask & \textbf{0.196} & 0.799 & 0.538 & 9.164 \\
\midrule
full+ T2I-Adapter+ w. mask & 0.188 & \textbf{0.822} & \textbf{0.496} & \textbf{9.631} \\
up+ T2I-Adapter+ w. mask & \textbf{0.193} & 0.778 & 0.552 & 9.099 \\
\midrule
full+ ControlNet+ w. mask & 0.189 & \textbf{0.818} & \textbf{0.504} & \textbf{9.560} \\
up+ ControlNet+ w. mask & \textbf{0.195} & 0.777 & 0.550 & 9.207 \\
\bottomrule
\end{tabular}
\end{adjustbox}
\end{table}

{\bf Mask-Guided Attention.}
\cref{tab:ablation_mask} analyzes the effect of mask guidance (\cref{fig:main_inject}). 
Applying mask during training and inference benefits appearance preservation.

\begin{table}[t]
\centering
\caption{Effect of mask-guided attention.}
\label{tab:ablation_mask}
\begin{adjustbox}{max width=\columnwidth}
\begin{tabular}{@{}lcccc@{}}
\toprule
Combination & CLIP-T $\uparrow$ & CLIP-I $\uparrow$ & LPIPS $\downarrow$ & PSNR $\uparrow$ \\
\midrule
full+ w/o disent.+ w. mask & \textbf{0.188} & \textbf{0.829} & \textbf{0.493} & \textbf{9.697} \\
full+ w/o disent.+ w/o mask & 0.183 & 0.784 & 0.522 & 9.488 \\
\midrule
up+ w/o disent.+ w. mask & \textbf{0.196} & \textbf{0.799} & \textbf{0.538} & \textbf{9.164} \\
up+ w/o disent.+ w/o mask & 0.191 & 0.770 & 0.544 & 9.137 \\
\bottomrule
\end{tabular}
\end{adjustbox}
\end{table}

\begin{table}[t]
\centering
\caption{Ablation study of controller-based layout/pose disentanglement.}
\label{tab:ablation_controller}
\begin{subtable}{\columnwidth}
  \centering
  \caption{Appearance preservation.}
  \label{tab:ablation_controller_app}
  \begin{adjustbox}{max width=\columnwidth}
  \begin{tabular}{@{}lcccc@{}}
  \toprule
  Combination & CLIP-T $\uparrow$ & CLIP-I $\uparrow$ & LPIPS $\downarrow$ & PSNR $\uparrow$ \\
  \midrule
  full+ w/o disent.+ w. mask & 0.188 & \textbf{0.829} & \textbf{0.493} & \textbf{9.697} \\
  full+ T2I-Adapter+ w. mask & 0.188 & 0.822 & 0.496 & 9.631 \\
  full+ ControlNet+ w. mask & \textbf{0.189} & 0.818 & 0.504 & 9.560 \\
  \bottomrule
  \end{tabular}
  \end{adjustbox}
\end{subtable}

\vspace{0.3em}
\begin{subtable}{\columnwidth}
  \centering
  \caption{Pose controllability.}
  \label{tab:ablation_controller_pose}
  \begin{adjustbox}{max width=\columnwidth}
  \begin{tabular}{@{}lccc@{}}
  \toprule
  Combination & AKD $\downarrow$ & MKR $\downarrow$ & Fail $\downarrow$ \\
  \midrule
  full+ w/o disent.+ w. mask & 0.112 & 0.109 & 0.242 \\
  full+ T2I-Adapter+ w. mask & \textbf{0.090} & \textbf{0.072} & \textbf{0.200} \\
  full+ ControlNet+ w. mask & 0.097 & 0.089 & 0.255 \\
  \bottomrule
  \end{tabular}
  \end{adjustbox}
\end{subtable}
\end{table}

{\bf Controller Disentanglement.}
\cref{tab:ablation_controller} compares T2I-Adapter and ControlNet, with qualitative results in \cref{fig:disen}. 
Using T2I-Adapter as disentanglement controller achieves lower AKD and MKR, suggesting more precise structural control.

{\bf CLIP Layer Selection.}
We ablate the number $k$ of aggregated CLIP layers in \cref{eq:clip-fuse}, varying $k\in\{3,7,11,15\}$ while fixing the full-injection${+}$T2I-Adapter${+}$mask configuration. As shown in \cref{tab:ablation_clip}, increasing $k$ steadily improves appearance preservation (higher CLIP-I, lower LPIPS, higher PSNR) and pose accuracy (lower AKD), since deeper aggregation supplies richer spatially aligned tokens. However, very large $k$ degrades text alignment (CLIP-T) and raises the pose-detection failure rate, as the model copies the reference layout more aggressively. We use $k{=}7$ in the main experiments as a balance between appearance fidelity, editability, and pose controllability.

\begin{table}[t]
\centering
\caption{Ablation on the number $k$ of aggregated CLIP layers.}
\label{tab:ablation_clip}
\begin{adjustbox}{max width=\columnwidth}
\begin{tabular}{@{}ccccccc@{}}
\toprule
$k$ & CLIP-T $\uparrow$ & CLIP-I $\uparrow$ & LPIPS $\downarrow$ & PSNR $\uparrow$ & AKD $\downarrow$ & MKR $\downarrow$ \\
\midrule
3  & 0.180 & 0.759 & 0.550 & 9.450 & 0.146 & 0.114 \\
7  & \textbf{0.188} & 0.822 & 0.496 & 9.631 & 0.090 & 0.072 \\
11 & 0.181 & 0.811 & 0.452 & 10.114 & 0.091 & \textbf{0.066} \\
15 & 0.175 & \textbf{0.831} & \textbf{0.359} & \textbf{11.207} & \textbf{0.068} & 0.083 \\
\bottomrule
\end{tabular}
\end{adjustbox}
\end{table}

\subsection{Application}

{\bf Compatibility with LoRA and External Conditions.}
Because AnimeAdapter only injects appearance features through the decoupled cross-attention and leaves the base SD weights untouched, it composes naturally with other plug-in modules. As shown in \cref{fig:application}, our method preserves character appearance while combining with additional concept conditions, structural controls, and different base models and LoRAs, supporting practical comic-style production workflows in which a consistent character is rendered across varied poses, scenes, and styles.
Beyond single-reference editing, our method also supports multi-subject driven generation via masked cross-attention (supplement, Multi-subject driven generation).

\section{Conclusion}
We propose AnimeAdapter, a lightweight adapter for consistent anime character generation under diverse editing conditions (\cref{fig:first}). First, by mapping and fusing CLIP's late-stage feature tokens, AnimeAdapter achieves fine-grained appearance preservation through U-Net decoupled cross-attention injection. We leverage CLIP emergent spatial localization capability to realize selective semantic appearance consistency. Second, we design a disentanglement training strategy that decouples appearance features from pose/layout information, enabling smooth and controllable character editing. Third, we present a high-quality dataset comprising curated anime character images with aligned OpenPose conditions, subject masks, and text prompts. We also introduce multi-dimensional evaluation protocols including text-guided local and global motion, facial expression, scene context, and OpenPose-conditioned editing. Extensive experiments on our evaluation tasks demonstrate strong fine-grained appearance preservation and editing capability (\cref{fig:comparison,fig:application}). To support reproducibility, our code, model weights, and dataset will be publicly released upon acceptance.

Future work will explore CLIP's interpretability in style transfer tasks and investigate the potential of diffusion-based models for real-time, lightweight anime generation in improvisational creative workflows.

\clearpage

{
    \small
    \bibliographystyle{ieeenat_fullname}
    \bibliography{main}

@String(CVPR  = {IEEE Conf. Comput. Vis. Pattern Recog.})

@String(ICLR  = {Int. Conf. Learn. Represent.})

@String(CVPR  = {CVPR})

@String(ICLR  = {ICLR})

@misc{IP-Adapter,
      title={IP-Adapter: Text Compatible Image Prompt Adapter for Text-to-Image Diffusion Models}, 
      author={Hu Ye and Jun Zhang and Sibo Liu and Xiao Han and Wei Yang},
      year={2023},
      eprint={2308.06721},
      archivePrefix={arXiv},
      primaryClass={cs.CV},
      url={https://arxiv.org/abs/2308.06721}, 
}

@misc{DreamBooth,
      title={DreamBooth: Fine Tuning Text-to-Image Diffusion Models for Subject-Driven Generation}, 
      author={Nataniel Ruiz and Yuanzhen Li and Varun Jampani and Yael Pritch and Michael Rubinstein and Kfir Aberman},
      year={2023},
      eprint={2208.12242},
      archivePrefix={arXiv},
      primaryClass={cs.CV},
      url={https://arxiv.org/abs/2208.12242}, 
}

@misc{T2I-Adapter,
      title={T2I-Adapter: Learning Adapters to Dig out More Controllable Ability for Text-to-Image Diffusion Models}, 
      author={Chong Mou and Xintao Wang and Liangbin Xie and Yanze Wu and Jian Zhang and Zhongang Qi and Ying Shan and Xiaohu Qie},
      year={2023},
      eprint={2302.08453},
      archivePrefix={arXiv},
      primaryClass={cs.CV},
      url={https://arxiv.org/abs/2302.08453}, 
}

@misc{InstantID,
      title={InstantID: Zero-shot Identity-Preserving Generation in Seconds}, 
      author={Qixun Wang and Xu Bai and Haofan Wang and Zekui Qin and Anthony Chen and Huaxia Li and Xu Tang and Yao Hu},
      year={2024},
      eprint={2401.07519},
      archivePrefix={arXiv},
      primaryClass={cs.CV},
      url={https://arxiv.org/abs/2401.07519}, 
}

@inproceedings{arar2023domain,
  title={Domain-agnostic tuning-encoder for fast personalization of text-to-image models},
  author={Arar, Moab and Gal, Rinon and Atzmon, Yuval and Chechik, Gal and Cohen-Or, Daniel and Shamir, Ariel and H. Bermano, Amit},
  booktitle={SIGGRAPH Asia 2023 Conference Papers},
  pages={1--10},
  year={2023}
}

@article{gu2023mix,
  title={Mix-of-show: Decentralized low-rank adaptation for multi-concept customization of diffusion models},
  author={Gu, Yuchao and Wang, Xintao and Wu, Jay Zhangjie and Shi, Yujun and Chen, Yunpeng and Fan, Zihan and Xiao, Wuyou and Zhao, Rui and Chang, Shuning and Wu, Weijia and others},
  journal={Advances in Neural Information Processing Systems},
  volume={36},
  pages={15890--15902},
  year={2023}
}

@article{hao2023vico,
  title={Vico: Plug-and-play visual condition for personalized text-to-image generation},
  author={Hao, Shaozhe and Han, Kai and Zhao, Shihao and Wong, Kwan-Yee K},
  journal={arXiv preprint arXiv:2306.00971},
  year={2023}
}

@inproceedings{ruiz2024hyperdreambooth,
  title={Hyperdreambooth: Hypernetworks for fast personalization of text-to-image models},
  author={Ruiz, Nataniel and Li, Yuanzhen and Jampani, Varun and Wei, Wei and Hou, Tingbo and Pritch, Yael and Wadhwa, Neal and Rubinstein, Michael and Aberman, Kfir},
  booktitle={Proceedings of the IEEE/CVF conference on computer vision and pattern recognition},
  pages={6527--6536},
  year={2024}
}

@article{smith2023continual,
  title={Continual diffusion: Continual customization of text-to-image diffusion with c-lora},
  author={Smith, James Seale and Hsu, Yen-Chang and Zhang, Lingyu and Hua, Ting and Kira, Zsolt and Shen, Yilin and Jin, Hongxia},
  journal={arXiv preprint arXiv:2304.06027},
  year={2023}
}

@misc{MasaCtrl,
      title={MasaCtrl: Tuning-Free Mutual Self-Attention Control for Consistent Image Synthesis and Editing}, 
      author={Mingdeng Cao and Xintao Wang and Zhongang Qi and Ying Shan and Xiaohu Qie and Yinqiang Zheng},
      year={2023},
      eprint={2304.08465},
      archivePrefix={arXiv},
      primaryClass={cs.CV},
      url={https://arxiv.org/abs/2304.08465}, 
}

@article{Dreamtuner,
  title={Dreamtuner: Single image is enough for subject-driven generation},
  author={Hua, Miao and Liu, Jiawei and Ding, Fei and Liu, Wei and Wu, Jie and He, Qian},
  journal={arXiv preprint arXiv:2312.13691},
  year={2023}
}

@inproceedings{Magicanimate,
  title={Magicanimate: Temporally consistent human image animation using diffusion model},
  author={Xu, Zhongcong and Zhang, Jianfeng and Liew, Jun Hao and Yan, Hanshu and Liu, Jia-Wei and Zhang, Chenxu and Feng, Jiashi and Shou, Mike Zheng},
  booktitle={Proceedings of the IEEE/CVF Conference on Computer Vision and Pattern Recognition},
  pages={1481--1490},
  year={2024}
}

@inproceedings{Animateanyone,
  title={Animateanyone: Consistent and controllable image-to-video synthesis for character animation},
  author={Hu, Li},
  booktitle={Proceedings of the IEEE/CVF Conference on Computer Vision and Pattern Recognition},
  pages={8153--8163},
  year={2024}
}

@inproceedings{Ssr-encoder,
  title={Ssr-encoder: Encoding selective subject representation for subject-driven generation},
  author={Zhang, Yuxuan and Song, Yiren and Liu, Jiaming and Wang, Rui and Yu, Jinpeng and Tang, Hao and Li, Huaxia and Tang, Xu and Hu, Yao and Pan, Han and others},
  booktitle={Proceedings of the IEEE/CVF Conference on Computer Vision and Pattern Recognition},
  pages={8069--8078},
  year={2024}
}

@inproceedings{Disco,
  title={Disco: Disentangled control for realistic human dance generation},
  author={Wang, Tan and Li, Linjie and Lin, Kevin and Zhai, Yuanhao and Lin, Chung-Ching and Yang, Zhengyuan and Zhang, Hanwang and Liu, Zicheng and Wang, Lijuan},
  booktitle={Proceedings of the IEEE/CVF Conference on Computer Vision and Pattern Recognition},
  pages={9326--9336},
  year={2024}
}

@inproceedings{han2024face,
  title={Face-adapter for pre-trained diffusion models with fine-grained id and attribute control},
  author={Han, Yue and Zhu, Junwei and He, Keke and Chen, Xu and Ge, Yanhao and Li, Wei and Li, Xiangtai and Zhang, Jiangning and Wang, Chengjie and Liu, Yong},
  booktitle={European Conference on Computer Vision},
  pages={20--36},
  year={2024},
  organization={Springer}
}

@inproceedings{ramesh2021zero,
  title={Zero-shot text-to-image generation},
  author={Ramesh, Aditya and Pavlov, Mikhail and Goh, Gabriel and Gray, Scott and Voss, Chelsea and Radford, Alec and Chen, Mark and Sutskever, Ilya},
  booktitle={International conference on machine learning},
  pages={8821--8831},
  year={2021},
  organization={Pmlr}
}

@article{ho2020denoising,
  title={Denoising diffusion probabilistic models},
  author={Ho, Jonathan and Jain, Ajay and Abbeel, Pieter},
  journal={Advances in neural information processing systems},
  volume={33},
  pages={6840--6851},
  year={2020}
}

@article{nichol2021glide,
  title={Glide: Towards photorealistic image generation and editing with text-guided diffusion models},
  author={Nichol, Alex and Dhariwal, Prafulla and Ramesh, Aditya and Shyam, Pranav and Mishkin, Pamela and McGrew, Bob and Sutskever, Ilya and Chen, Mark},
  journal={arXiv preprint arXiv:2112.10741},
  year={2021}
}

@article{ho2022classifier,
  title={Classifier-free diffusion guidance},
  author={Ho, Jonathan and Salimans, Tim},
  journal={arXiv preprint arXiv:2207.12598},
  year={2022}
}

@article{saharia2022photorealistic,
  title={Photorealistic text-to-image diffusion models with deep language understanding},
  author={Saharia, Chitwan and Chan, William and Saxena, Saurabh and Li, Lala and Whang, Jay and Denton, Emily L and Ghasemipour, Kamyar and Gontijo Lopes, Raphael and Karagol Ayan, Burcu and Salimans, Tim and others},
  journal={Advances in neural information processing systems},
  volume={35},
  pages={36479--36494},
  year={2022}
}

@article{ramesh2022hierarchical,
  title={Hierarchical text-conditional image generation with clip latents},
  author={Ramesh, Aditya and Dhariwal, Prafulla and Nichol, Alex and Chu, Casey and Chen, Mark},
  journal={arXiv preprint arXiv:2204.06125},
  volume={1},
  number={2},
  pages={3},
  year={2022}
}

@inproceedings{rombach2022high,
  title={High-resolution image synthesis with latent diffusion models},
  author={Rombach, Robin and Blattmann, Andreas and Lorenz, Dominik and Esser, Patrick and Ommer, Bj{\"o}rn},
  booktitle={Proceedings of the IEEE/CVF conference on computer vision and pattern recognition},
  pages={10684--10695},
  year={2022}
}

@inproceedings{peebles2023scalable,
  title={Scalable diffusion models with transformers},
  author={Peebles, William and Xie, Saining},
  booktitle={Proceedings of the IEEE/CVF international conference on computer vision},
  pages={4195--4205},
  year={2023}
}

@inproceedings{zhang2023adding,
  title={Adding conditional control to text-to-image diffusion models},
  author={Zhang, Lvmin and Rao, Anyi and Agrawala, Maneesh},
  booktitle={Proceedings of the IEEE/CVF international conference on computer vision},
  pages={3836--3847},
  year={2023}
}

@inproceedings{li2023gligen,
  title={Gligen: Open-set grounded text-to-image generation},
  author={Li, Yuheng and Liu, Haotian and Wu, Qingyang and Mu, Fangzhou and Yang, Jianwei and Gao, Jianfeng and Li, Chunyuan and Lee, Yong Jae},
  booktitle={Proceedings of the IEEE/CVF conference on computer vision and pattern recognition},
  pages={22511--22521},
  year={2023}
}

@inproceedings{xie2023boxdiff,
  title={Boxdiff: Text-to-image synthesis with training-free box-constrained diffusion},
  author={Xie, Jinheng and Li, Yuexiang and Huang, Yawen and Liu, Haozhe and Zhang, Wentian and Zheng, Yefeng and Shou, Mike Zheng},
  booktitle={Proceedings of the IEEE/CVF International Conference on Computer Vision},
  pages={7452--7461},
  year={2023}
}

@inproceedings{shi2024dragdiffusion,
  title={Dragdiffusion: Harnessing diffusion models for interactive point-based image editing},
  author={Shi, Yujun and Xue, Chuhui and Liew, Jun Hao and Pan, Jiachun and Yan, Hanshu and Zhang, Wenqing and Tan, Vincent YF and Bai, Song},
  booktitle={Proceedings of the IEEE/CVF Conference on Computer Vision and Pattern Recognition},
  pages={8839--8849},
  year={2024}
}

@article{hu2022lora,
  title={Lora: Low-rank adaptation of large language models.},
  author={Hu, Edward J and Shen, Yelong and Wallis, Phillip and Allen-Zhu, Zeyuan and Li, Yuanzhi and Wang, Shean and Wang, Lu and Chen, Weizhu and others},
  journal={ICLR},
  volume={1},
  number={2},
  pages={3},
  year={2022}
}

@article{zhou2024storydiffusion,
  title={Storydiffusion: Consistent self-attention for long-range image and video generation},
  author={Zhou, Yupeng and Zhou, Daquan and Cheng, Ming-Ming and Feng, Jiashi and Hou, Qibin},
  journal={Advances in Neural Information Processing Systems},
  volume={37},
  pages={110315--110340},
  year={2024}
}

@inproceedings{chen2024anydoor,
  title={Anydoor: Zero-shot object-level image customization},
  author={Chen, Xi and Huang, Lianghua and Liu, Yu and Shen, Yujun and Zhao, Deli and Zhao, Hengshuang},
  booktitle={Proceedings of the IEEE/CVF conference on computer vision and pattern recognition},
  pages={6593--6602},
  year={2024}
}

@inproceedings{patashnik2024consolidating,
  title={Consolidating attention features for multi-view image editing},
  author={Patashnik, Or and Gal, Rinon and Cohen-Or, Daniel and Zhu, Jun-Yan and De la Torre, Fernando},
  booktitle={SIGGRAPH Asia 2024 Conference Papers},
  pages={1--12},
  year={2024}
}

@article{shen2023advancing,
  title={Advancing pose-guided image synthesis with progressive conditional diffusion models},
  author={Shen, Fei and Ye, Hu and Zhang, Jun and Wang, Cong and Han, Xiao and Yang, Wei},
  journal={arXiv preprint arXiv:2310.06313},
  year={2023}
}

@article{ai2023stable,
  title={Stable diffusion reference only: Image prompt and blueprint jointly guided multi-condition diffusion model for secondary painting},
  author={Ai, Hao and Sheng, Lu},
  journal={arXiv preprint arXiv:2311.02343},
  year={2023}
}

@article{song2020denoising,
  title={Denoising diffusion implicit models},
  author={Song, Jiaming and Meng, Chenlin and Ermon, Stefano},
  journal={arXiv preprint arXiv:2010.02502},
  year={2020}
}

@article{lu2022dpm,
  title={Dpm-solver: A fast ode solver for diffusion probabilistic model sampling in around 10 steps},
  author={Lu, Cheng and Zhou, Yuhao and Bao, Fan and Chen, Jianfei and Li, Chongxuan and Zhu, Jun},
  journal={Advances in neural information processing systems},
  volume={35},
  pages={5775--5787},
  year={2022}
}

@article{liu2022pseudo,
  title={Pseudo numerical methods for diffusion models on manifolds},
  author={Liu, Luping and Ren, Yi and Lin, Zhijie and Zhao, Zhou},
  journal={arXiv preprint arXiv:2202.09778},
  year={2022}
}

@article{gandelsman2023interpreting,
  title={Interpreting clip's image representation via text-based decomposition},
  author={Gandelsman, Yossi and Efros, Alexei A and Steinhardt, Jacob},
  journal={arXiv preprint arXiv:2310.05916},
  year={2023}
}

@misc{SAM3,
      title={SAM 3: Segment Anything with Concepts},
      author={Nicolas Carion and Laura Gustafson and Yuan-Ting Hu and Shoubhik Debnath and Ronghang Hu and Didac Suris and Chaitanya Ryali and Kalyan Vasudev Alwala and Haitham Khedr and Andrew Huang and Jie Lei and Tengyu Ma and Baishan Guo and Arpit Kalla and Markus Marks and Joseph Greer and Meng Wang and Peize Sun and Roman Rädle and Triantafyllos Afouras and Effrosyni Mavroudi and Katherine Xu and Tsung-Han Wu and Yu Zhou and Liliane Momeni and Rishi Hazra and Shuangrui Ding and Sagar Vaze and Francois Porcher and Feng Li and Siyuan Li and Aishwarya Kamath and Ho Kei Cheng and Piotr Dollár and Nikhila Ravi and Kate Saenko and Pengchuan Zhang and Christoph Feichtenhofer},
      year={2025},
      eprint={2511.16719},
      archivePrefix={arXiv},
      primaryClass={cs.CV},
      url={https://arxiv.org/abs/2511.16719},
}

@article{clip_i_cc_2024,
  title={AutoStudio: Crafting Consistent Subjects in Multi-turn Interactive Image Generation},
  author={Cheng, Junhao and Lu, Xi and Li, Hanhui and Zai, Khun Loun and Yin, Baiqiao and Cheng, Yuhao and Yan, Yiqiang and Liang, Xiaodan},
  journal={arXiv preprint arXiv:2406.01388},
  year={2024}
}

@inproceedings{hessel2021clipscore,
  title={CLIPScore: A Reference-free Evaluation Metric for Image Captioning},
  author={Hessel, Jack and Holtzman, Ari and Forbes, Maxwell and Le Bras, Ronan and Choi, Yejin},
  booktitle={EMNLP},
  year={2021}
}

@inproceedings{zhang2018lpips,
  title={The Unreasonable Effectiveness of Deep Features as a Perceptual Metric},
  author={Zhang, Richard and Isola, Phillip and Efros, Alexei A and Shechtman, Eli and Wang, Oliver},
  booktitle={CVPR},
  year={2018}
}

@article{instantstyle,
  title={Instantstyle: Free lunch towards style-preserving in text-to-image generation},
  author={Wang, Haofan and Spinelli, Matteo and Wang, Qixun and Bai, Xu and Qin, Zekui and Chen, Anthony},
  journal={arXiv preprint arXiv:2404.02733},
  year={2024}
}

@article{textualinversion,
  title={An image is worth one word: Personalizing text-to-image generation using textual inversion},
  author={Gal, Rinon and Alaluf, Yuval and Atzmon, Yuval and Patashnik, Or and Bermano, Amit H and Chechik, Gal and Cohen-Or, Daniel},
  journal={arXiv preprint arXiv:2208.01618},
  year={2022}
}

@misc{animaginexl4,
  title={Animagine XL 4.0},
  author={{Cagliostro Research Lab}},
  year={2025},
  howpublished={\url{https://huggingface.co/cagliostrolab/animagine-xl-4.0}},
  note={Anime-themed text-to-image diffusion model fine-tuned from Stable Diffusion XL 1.0}
}

@misc{danbooru2025metadata,
  title={Danbooru 2025 Metadata},
  author={trojblue},
  year={2025},
  howpublished={\url{https://huggingface.co/datasets/trojblue/danbooru2025-metadata}},
  note={Structured metadata for Danbooru posts in Parquet format}
}

@inproceedings{peng2025iclr-dreambench,
  title={{DreamBench++}: A Human-Aligned Benchmark for Personalized Image Generation},
  author={Peng, Yuang and Cui, Yuxin and Tang, Haomiao and Qi, Zekun and Dong, Runpei and Bai, Jing and Han, Chunrui and Ge, Zheng and Zhang, Xiangyu and Xia, Shu-Tao},
  booktitle=ICLR,
  year={2025}
}
}

\end{document}

\title{AnimeAdapter: A Modular Adapter for Appearance-Consistent Anime Character Generation}

\maketitlesupplementary

\appendix

\section{Preliminary}
Diffusion models consist of a forward diffusion process and a reverse sampling process. In the forward process, Gaussian noise is progressively added to the data sample $x_0$, resulting in a sequence of noisy variables $\{x_t\}_{t=1}^T$. The transition at each timestep is defined as
\begin{equation}
x_t = \sqrt{\alpha_t} x_{t-1} + \sqrt{\beta_t} \epsilon_t,
\end{equation}
where $\epsilon_t \sim \mathcal{N}(0, I)$, $\beta_t = 1 - \alpha_t$, and $\{\alpha_t\}$ is a monotonically decreasing sequence. The timestep $t \in \{1,\dots,T\}$ controls the noise level, and $T$ is chosen sufficiently
large such that $x_T$ approximates an isotropic Gaussian distribution. Equation~(1) defines the forward transition distribution $q(x_t \mid x_{t-1})$. By recursively expanding the forward process, a closed-form expression can be obtained:
\begin{equation}
x_t = \sqrt{\bar{\alpha}_t} x_0 + \sqrt{1 - \bar{\alpha}_t} \epsilon,
\end{equation}
where $\bar{\alpha}_t = \prod_{i=1}^t \alpha_i$ and $\epsilon \sim \mathcal{N}(0, I)$. This formulation characterizes the marginal distribution $q(x_t \mid x_0)$. The sampling process corresponds to the reverse diffusion procedure, which aims to recover $x_0$ by iteratively denoising a Gaussian noise sample. In the case of DDPM~\cite{ho2020denoising}, sampling requires the reverse transition distribution $p_\theta(x_{t-1} \mid x_t)$. Since this distribution is intractable, the posterior $q(x_{t-1} \mid x_t, x_0)$ of the forward Markov chain is first derived using Bayes' rule, yielding
\begin{equation}
x_{t-1} = \tilde{\mu}(x_0, x_t) + \sigma_t z,
\end{equation}
where $z \sim \mathcal{N}(0, I)$, $\sigma_t$ is a predefined variance, and $\tilde{\mu}(\cdot)$ is a deterministic function with fixed coefficients. By substituting the expression of $x_0$ derived from Equation~(2), the mean term
$\tilde{\mu}(x_0, x_t)$ can be reparameterized in terms of $\epsilon$ and $x_t$. Consequently, the learning objective can be simplified to training a neural network $\epsilon_\theta$ to predict the noise $\epsilon$ from $x_t$ and the timestep $t$. The training objective is defined as
\begin{equation}
\mathcal{L}
= \mathbb{E}\big[\|\epsilon - \epsilon_\theta(x_t, c, t)\|^2\big],
\end{equation}
where $c$ denotes optional conditioning information. As a variant, the diffusion prior in DALL·E~2~\cite{ramesh2022hierarchical} directly parameterizes the reverse process by predicting $x_{t-1}$ instead of $\epsilon$. In addition, several accelerated sampling methods, including DDIM~\cite{song2020denoising}, DPM-Solver~\cite{lu2022dpm}, and PNDM~\cite{liu2022pseudo}, have been proposed to reduce the number of sampling steps while preserving generation quality.

Classifier-free guidance~\cite{ho2022classifier} enhances conditional controllability without requiring an external classifier. During training, conditioning information is randomly dropped with a fixed probability. At inference time, the guided noise prediction is computed as a linear combination of conditional and unconditional predictions:
\begin{equation}
\dot{\epsilon}
= w \, \epsilon_\theta(x_t, c, t)
+ (1 - w)\, \epsilon_\theta(x_t, t),
\label{eq:cfg_epsilon}
\end{equation}
where $w$ controls the guidance strength.

\section{Detailed Evaluation Dataset Construction}
\label{sec:supp_eval_dataset}
This section provides the complete implementation details of evaluation dataset construction.

\subsection{Semantic Cluster Extraction}

Given a Danbooru metadata entry, we first extract semantic clusters:

\begin{equation}
S =
\{
C_0, C_1, C_2, C_3, C_4, C_5
\}
\end{equation}

where

\begin{itemize}
\item $C_0$: identity and character count
\item $C_1$: view and framing
\item $C_2$: style and aesthetic
\item $C_3$: posture
\item $C_4$: fine-grained body-part attributes
\item $C_5$: patterns
\end{itemize}

\subsection{Reference Prompt Construction}

Multiple reference prompts are constructed for different evaluation tasks.

\paragraph{Original reference prompt}
\begin{equation}
\begin{aligned}
P_{ref}^{orig} =\;& C_0 + C_{char\_name} + rating \\
&+ C_1 + C_2 + C_3 + C_4 + quality
\end{aligned}
\end{equation}
where $C_{char\_name}$ denotes the character identity tags, and $quality$ denotes aesthetic quality tokens including \texttt{masterpiece}, \texttt{high score}, \texttt{great score}, and \texttt{absurdres}.

\paragraph{Full-body reference prompt}
\begin{equation}
\begin{aligned}
P_{ref}^{full} =\;& C_0 + C_{char\_name} + \texttt{full-body} \\
&+ neutral\_view + C_2 + quality
\end{aligned}
\end{equation}
Here, $neutral\_view$ includes \texttt{straight-on} and \texttt{looking at viewer}, which reduce viewpoint bias and provide consistent structural baselines.

\paragraph{Upper-body reference prompt}
\begin{equation}
\begin{aligned}
P_{ref}^{upper} =\;& C_0 + C_{char\_name} + \texttt{upper-body} \\
&+ neutral\_view + C_2 + quality
\end{aligned}
\end{equation}
Additionally, we remove samples in which \texttt{full-body} or \texttt{upper-body} conflicts with $C_2$ tags.

\paragraph{Portrait reference prompt}
\begin{equation}
\begin{aligned}
P_{ref}^{portrait} =\;& C_0 + C_{char\_name} \\
&+ \texttt{portrait} + C_2 + quality
\end{aligned}
\end{equation}

Reference images are generated from these prompts.

\subsection{Editing Prompt Construction}

Editing prompts are constructed using structured transformations.

\paragraph{Body-part motion editing}
Body-part motion tags are substituted:
\begin{equation}
P_{edit}^{appearance} = C_0 + substitute(C_4) + quality
\end{equation}
When substituting $C_4$ tags, we identify the corresponding semantic sub-group and randomly sample a replacement local motion tag from the same sub-group.

\paragraph{Text-driven posture and viewpoint editing}
Pose and framing tags are modified:
\begin{equation}
\begin{aligned}
P_{edit}^{pose} =\;& C_0 + \texttt{full-body or upper-body} \\
&+ C_1 + new\_pose + quality
\end{aligned}
\end{equation}

\paragraph{Facial expression editing}
Facial motion attributes are introduced:
\begin{equation}
P_{edit}^{expression} = C_0 + \texttt{portrait} + new\_expression + quality
\end{equation}

\paragraph{Scene editing}
Scene descriptions are introduced:
\begin{equation}
P_{edit}^{scene} = C_0 + scene\_description
\end{equation}

where $new\_pose$ and $new\_expression$ are derived from the reorganized tag groups, and $scene\_description$ is a set of scene-related prompts created by us.

\subsection{Pose-conditioned Editing}

An OpenPose skeleton extracted from the reference image is used as the conditioning signal:

\begin{equation}
I_{edit}
=
Model(I_{ref}, P_{edit}, Pose(I_{ref}))
\end{equation}

\subsection{Example}

We provide a concrete example illustrating how prompt building components are constructed from Danbooru metadata.

\paragraph{Original Danbooru tags.}
\begin{quote}
\small\sloppy\raggedright
\texttt{tag\_string\_general:}\\
1girl, solo, abstract, biopunk, bike shorts, blonde hair, blouse, blue eyes, cosplay, from below, cowboy shot, double bun, from behind, gloves, hair bun, hair rings, long hair, looking at viewer, pleated skirt, purple skirt, school uniform, shirt, short sleeves, shorts, shorts under skirt, skirt, unworn shorts, vest, white background, white shirt

\texttt{tag\_string\_character:}\\
yuugumo (kancolle), abukuma (kancolle), yuugumo kai ni (kancolle)

\texttt{tag\_string\_artist:}\\
Channel (Caststation)

\texttt{rating:} general
\end{quote}

\paragraph{Extracted prompt components.}
After semantic clustering and preprocessing, the prompt building components are:
\begin{quote}
\small\sloppy\raggedright

\textbf{Character identity:}
$C_{char\_name}$ = \{
yuugumo (kancolle),
abukuma (kancolle),
yuugumo kai ni (kancolle)
\}

\textbf{Cluster 0 (identity and count):}

$C_0$ =
\{
1girl, solo
\}

\textbf{Cluster 1 (view and framing):}

$C_1$ =
\{
from behind, from below, cowboy shot
\}

\textbf{Cluster 2 (style and aesthetics):}

$C_2$ =
\{
abstract,
biopunk,
Channel (Caststation),
simple background,
white background
\}

\textbf{Cluster 3 (posture):}

$C_3$ = \{\}

\textbf{Cluster 4 (body-part attributes):}

$C_4$ =
\{
hair rings,
hair bun,
hair flip,
double bun,
blonde hair,
long hair,
white shirt,
shirt,
blouse,
school uniform,
gloves,
short sleeves,
vest,
blue eyes,
eye pop,
pleated skirt,
bike shorts,
shorts,
skirt,
shorts under skirt,
purple skirt,
unworn shorts
\}

\textbf{Quality:}

quality =
\{
masterpiece,
great score,
high score,
absurdres
\}

\end{quote}

\paragraph{Edited components used in evaluation.}
During evaluation prompt construction, selected components are substituted or introduced:
\begin{quote}
\small\sloppy\raggedright

\textbf{Substituted Cluster 4:}

substituted($C_4$) =
\{
eating hair,
> <
hair rings,
hair bun,
double bun,
blonde hair,
long hair,
white shirt,
shirt,
blouse,
school uniform,
gloves,
short sleeves,
vest,
blue eyes,
pleated skirt,
bike shorts,
shorts,
skirt,
shorts under skirt,
purple skirt,
unworn shorts
\}
(replacing hair flip and eye pop)

\textbf{Neutral view tokens:}

neutral\_view =
\{
straight-on,
looking at viewer
\}

\textbf{New pose attribute:}

new\_pose =
\{
wing hug
\}

\textbf{New expression attribute:}

new\_expression =
\{
furrowed brow
\}

\textbf{Scene description:}

scene\_description =
\{
in the kitchen
\}

\end{quote}

\section{Total Quantitative Results}
\label{sec:supp_quantitative}

\cref{tab:body_edit,tab:fullbody_posture,tab:upperbody_posture,tab:portrait_expression,tab:scenario,tab:pose_conditioned_fullbody} show the complete quantitative evaluation results across all evaluation tasks.

\begin{table}[t]
\centering
\caption{Quantitative comparison on Body-part Motion Editing.}
\begin{adjustbox}{max width=\columnwidth}
\footnotesize
\setlength{\tabcolsep}{2pt}
\begin{tabular}{@{}lccccc@{}}
\toprule
Method & CLIP-T $\uparrow$ & CLIP-I $\uparrow$ & LPIPS $\downarrow$ & LPIPS-div $\uparrow$ & PSNR $\uparrow$ \\
\midrule
full w/o disent. w. mask & 0.189 & 0.776 & 0.564 & 0.522 & 9.057 \\
full w/o disent. w/o mask & 0.186 & 0.745 & 0.589 & \textbf{0.557} & 9.011 \\
up w/o disent. w. mask & 0.194 & 0.769 & 0.582 & 0.526 & 8.689 \\
up w/o disent. w/o mask & 0.193 & 0.741 & 0.589 & 0.535 & 8.856 \\
full + T2I-Adapter & 0.189 & 0.750 & 0.559 & 0.527 & 9.095 \\
full + ControlNet & 0.191 & 0.763 & 0.555 & 0.522 & 9.178 \\
up + ControlNet & \textbf{0.194} & 0.748 & 0.591 & 0.540 & 8.886 \\
up + T2I-Adapter & 0.193 & 0.740 & 0.589 & 0.530 & 8.849 \\
IP-Adapter & 0.187 & 0.745 & 0.550 & 0.524 & 10.654 \\
IP-Adapter BigG & 0.187 & 0.736 & 0.553 & 0.511 & 10.611 \\
IP-Adapter Plus & 0.183 & 0.766 & \textbf{0.487} & 0.444 & 12.269 \\
Ours (Full+Mask+Ctrl) & 0.187 & \textbf{0.799} & 0.502 & 0.496 & \textbf{12.608} \\
\bottomrule
\end{tabular}
\end{adjustbox}
\label{tab:body_edit}
\end{table}
\begin{table}[t]
\centering
\caption{Quantitative comparison on Upper-body Posture Editing.}
\begin{adjustbox}{max width=\columnwidth}
\footnotesize
\setlength{\tabcolsep}{2pt}
\begin{tabular}{@{}lccccc@{}}
\toprule
Method & CLIP-T $\uparrow$ & CLIP-I $\uparrow$ & LPIPS $\downarrow$ & LPIPS-div $\uparrow$ & PSNR $\uparrow$ \\
\midrule
full w/o disent. w. mask & 0.154 & 0.862 & 0.538 & 0.497 & 8.431 \\
full w/o disent. w/o mask & 0.153 & 0.849 & 0.527 & 0.499 & 8.509 \\
up w/o disent. w. mask & 0.166 & 0.842 & 0.580 & 0.502 & 7.991 \\
up w/o disent. w/o mask & 0.160 & 0.821 & 0.575 & 0.515 & 8.105 \\
full + T2I-Adapter & 0.156 & \textbf{0.868} & 0.556 & \textbf{0.527} & 8.453 \\
full + ControlNet & 0.159 & 0.860 & 0.567 & 0.527 & 8.370 \\
up + ControlNet & \textbf{0.166} & 0.813 & 0.604 & 0.521 & 7.888 \\
up + T2I-Adapter & 0.164 & 0.813 & 0.606 & 0.509 & 7.825 \\
IP-Adapter & 0.154 & 0.801 & 0.568 & 0.511 & 8.466 \\
IP-Adapter BigG & 0.151 & 0.801 & 0.562 & 0.481 & 8.582 \\
IP-Adapter Plus & 0.152 & 0.845 & 0.497 & 0.460 & 9.634 \\
Ours (Full+Mask+Ctrl) & 0.157 & 0.866 & \textbf{0.486} & 0.485 & \textbf{9.962} \\
\bottomrule
\end{tabular}
\end{adjustbox}
\label{tab:upperbody_posture}
\end{table}
\begin{table}[t]
\centering
\caption{Quantitative comparison on Text-driven Posture and Viewpoint Editing.}
\begin{adjustbox}{max width=\columnwidth}
\footnotesize
\setlength{\tabcolsep}{2pt}
\begin{tabular}{@{}lccccc@{}}
\toprule
Method & CLIP-T $\uparrow$ & CLIP-I $\uparrow$ & LPIPS $\downarrow$ & LPIPS-div $\uparrow$ & PSNR $\uparrow$ \\
\midrule
full w/o disent. w. mask & 0.181 & 0.839 & 0.414 & 0.396 & 10.955 \\
full w/o disent. w/o mask & 0.177 & 0.751 & 0.485 & 0.461 & 10.286 \\
up w/o disent. w. mask & \textbf{0.194} & 0.799 & 0.461 & 0.435 & 10.533 \\
up w/o disent. w/o mask & 0.183 & 0.730 & 0.490 & \textbf{0.476} & 10.098 \\
full + T2I-Adapter & 0.181 & 0.841 & 0.413 & 0.447 & 10.669 \\
full + ControlNet & 0.181 & 0.837 & 0.426 & 0.432 & 10.597 \\
up + ControlNet & 0.191 & 0.786 & 0.471 & 0.454 & 10.453 \\
up + T2I-Adapter & 0.191 & 0.796 & 0.471 & 0.456 & 10.291 \\
IP-Adapter & 0.177 & 0.801 & 0.437 & 0.408 & 11.222 \\
IP-Adapter BigG & 0.178 & 0.793 & 0.457 & 0.424 & 10.821 \\
IP-Adapter Plus & 0.173 & 0.815 & 0.375 & 0.392 & \textbf{12.032} \\
Ours (Full+Mask+Ctrl) & 0.181 & \textbf{0.870} & \textbf{0.357} & 0.391 & 11.314 \\
\bottomrule
\end{tabular}
\end{adjustbox}
\label{tab:fullbody_posture}
\end{table}
\begin{table}[t]
\centering
\caption{Quantitative comparison on Facial Expression Editing.}
\begin{adjustbox}{max width=\columnwidth}
\footnotesize
\setlength{\tabcolsep}{2pt}
\begin{tabular}{@{}lccccc@{}}
\toprule
Method & CLIP-T $\uparrow$ & CLIP-I $\uparrow$ & LPIPS $\downarrow$ & LPIPS-div $\uparrow$ & PSNR $\uparrow$ \\
\midrule
full w/o disent. w. mask & 0.178 & 0.866 & 0.521 & 0.501 & 8.464 \\
full w/o disent. w/o mask & 0.179 & 0.864 & 0.511 & 0.490 & 8.578 \\
up w/o disent. w. mask & 0.183 & 0.861 & 0.546 & 0.491 & 8.073 \\
up w/o disent. w/o mask & 0.183 & 0.850 & 0.539 & 0.494 & 8.217 \\
full + T2I-Adapter & 0.179 & 0.851 & 0.524 & \textbf{0.521} & 8.485 \\
full + ControlNet & 0.182 & 0.854 & 0.540 & 0.515 & 8.377 \\
up + ControlNet & \textbf{0.184} & 0.839 & 0.565 & 0.511 & 8.104 \\
up + T2I-Adapter & 0.181 & 0.833 & 0.574 & 0.505 & 7.989 \\
IP-Adapter & 0.175 & 0.828 & 0.548 & 0.499 & 8.683 \\
IP-Adapter BigG & 0.172 & 0.823 & 0.546 & 0.465 & 8.638 \\
IP-Adapter Plus & 0.178 & 0.854 & 0.478 & 0.433 & 9.848 \\
Ours (Full+Mask+Ctrl) & 0.180 & \textbf{0.895} & \textbf{0.455} & 0.474 & \textbf{10.193} \\
\bottomrule
\end{tabular}
\end{adjustbox}
\label{tab:portrait_expression}
\end{table}
\begin{table}[t]
\centering
\caption{Quantitative comparison on Scene Editing.}
\begin{adjustbox}{max width=\columnwidth}
\footnotesize
\setlength{\tabcolsep}{2pt}
\begin{tabular}{@{}lccccc@{}}
\toprule
Method & CLIP-T $\uparrow$ & CLIP-I $\uparrow$ & LPIPS $\downarrow$ & LPIPS-div $\uparrow$ & PSNR $\uparrow$ \\
\midrule
full w/o disent. w. mask & 0.234 & 0.848 & 0.451 & 0.440 & 10.539 \\
full w/o disent. w/o mask & 0.216 & 0.813 & 0.492 & 0.495 & 10.048 \\
up w/o disent. w. mask & \textbf{0.242} & 0.810 & 0.498 & 0.486 & 10.064 \\
up w/o disent. w/o mask & 0.237 & 0.807 & 0.504 & 0.493 & 9.764 \\
full + T2I-Adapter & 0.232 & 0.837 & 0.452 & 0.465 & 10.310 \\
full + ControlNet & 0.231 & 0.821 & 0.470 & 0.457 & 10.261 \\
up + ControlNet & 0.240 & 0.809 & 0.506 & \textbf{0.496} & 9.941 \\
up + T2I-Adapter & 0.237 & 0.809 & 0.507 & 0.492 & 9.838 \\
IP-Adapter & 0.194 & 0.802 & 0.439 & 0.428 & 11.059 \\
IP-Adapter BigG & 0.191 & 0.782 & 0.456 & 0.440 & 10.771 \\
IP-Adapter Plus & 0.179 & 0.814 & 0.384 & 0.385 & 11.804 \\
Ours (Full+Mask+Ctrl) & 0.224 & \textbf{0.881} & \textbf{0.375} & 0.417 & \textbf{12.005} \\
\bottomrule
\end{tabular}
\end{adjustbox}
\label{tab:scenario}
\end{table}
\begin{table}[t]
\centering
\caption{Quantitative comparison on Pose-conditioned Editing.}
\begin{adjustbox}{max width=\columnwidth}
\footnotesize
\setlength{\tabcolsep}{1.5pt}
\begin{tabular}{@{}lcccccccc@{}}
\toprule
Method & CLIP-T $\uparrow$ & CLIP-I $\uparrow$ & LPIPS $\downarrow$ & LPIPS-div $\uparrow$ & PSNR $\uparrow$ & AKD $\downarrow$ & MKR $\downarrow$ & Failure $\downarrow$ \\
\midrule
full w/o disent. w. mask & 0.192 & 0.784 & 0.473 & 0.483 & 10.739 & 0.112 & 0.109 & 0.242 \\
full w/o disent. w/o mask & 0.185 & 0.683 & 0.530 & 0.487 & 10.498 & 0.115 & 0.102 & 0.258 \\
up w/o disent. w. mask & \textbf{0.197} & 0.715 & 0.561 & \textbf{0.572} & 9.635 & 0.192 & 0.126 & 0.448 \\
up w/o disent. w/o mask & 0.190 & 0.668 & 0.570 & 0.562 & 9.780 & 0.177 & 0.117 & 0.417 \\
full + T2I-Adapter & 0.191 & 0.787 & 0.474 & 0.479 & 10.774 & 0.090 & 0.072 & \textbf{0.200} \\
full + ControlNet & 0.189 & 0.774 & 0.467 & 0.507 & 10.580 & 0.097 & 0.089 & 0.255 \\
up + ControlNet & 0.193 & 0.669 & 0.564 & 0.560 & 9.971 & 0.191 & 0.139 & 0.417 \\
up + T2I-Adapter & 0.190 & 0.679 & 0.565 & 0.563 & 9.802 & 0.182 & 0.132 & 0.425 \\
IP-Adapter & 0.189 & 0.770 & 0.477 & 0.470 & 10.956 & 0.109 & 0.106 & 0.550 \\
IP-Adapter BigG & 0.187 & 0.773 & 0.476 & 0.452 & 11.181 & 0.104 & 0.098 & 0.310 \\
IP-Adapter Plus & 0.186 & 0.796 & \textbf{0.383} & 0.397 & 12.142 & \textbf{0.070} & 0.075 & 0.273 \\
Ours (Full+Mask+Ctrl) & 0.195 & \textbf{0.852} & 0.411 & 0.497 & \textbf{12.543} & 0.090 & \textbf{0.071} & 0.255 \\
\bottomrule
\end{tabular}
\end{adjustbox}
\label{tab:pose_conditioned_fullbody}
\end{table}

\section{Training CFG Setting}
\label{sec:supp_cfg}
To implement CFG for text and fine-grained image conditions, in each training step we sample a value $c$
from the interval $[0, 0.25)$ using a uniform distribution. When $c < 0.15$, we drop the reference image
embedding by setting $I_r$ to zeros. When $0.15 \leq c < 0.2$, we drop the text condition by providing an empty
string to the tokenizer. When $0.2 \leq c < 0.25$, we drop both conditions. We do not randomly discard the pose
condition when training with the frozen pose controller.

\begin{figure}[t]
  \centering
  \includegraphics[width=\columnwidth]{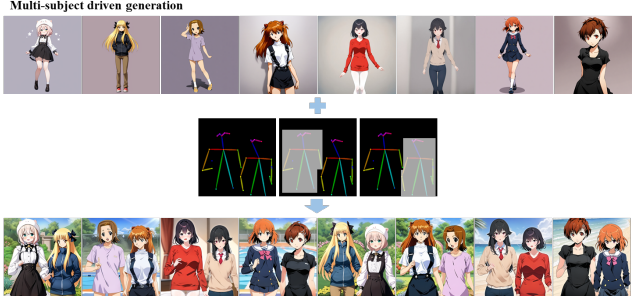}
  \caption{Multi-subject driven generation using our proposed method.}
  \label{fig:multi}
\end{figure}

\section{Multi-subject driven generation}
\label{sec:supp_multi}
As shown in \cref{fig:multi}, our method also works for multi-subject driven generation. The specific implementation is to use OpenPose conditions and apply masks on cross-attention maps.

\section{Failed OpenPose-conditioning Cases and Masking Robustness}
\label{sec:supp_fail_mask}
\begin{figure}[t]
  \centering
  \scriptsize
  \setlength{\tabcolsep}{1pt}
  \begin{tabular}{@{}cccc@{}}
    \FailMaskImg{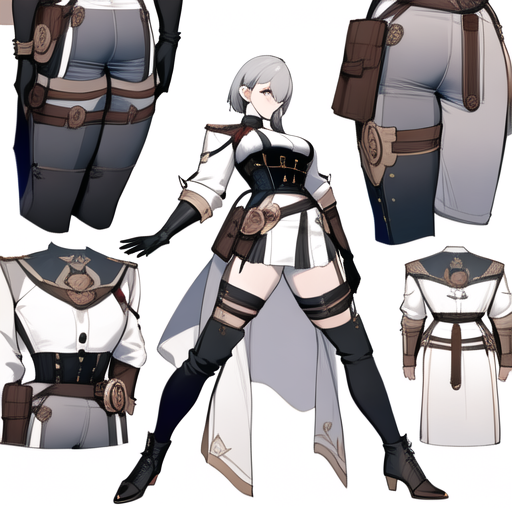}{\failimgw} &
    \FailMaskImg{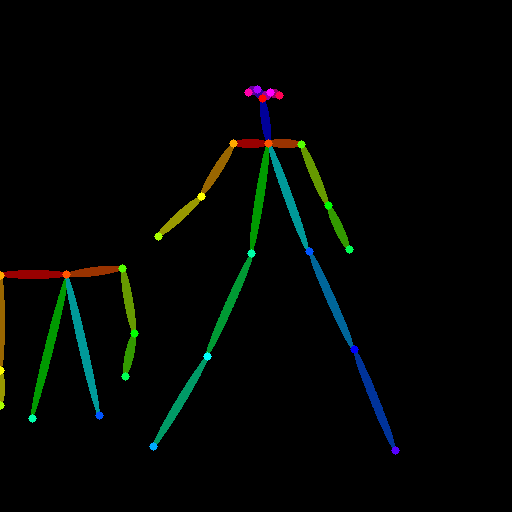}{\failimgw} &
    \FailMaskImg{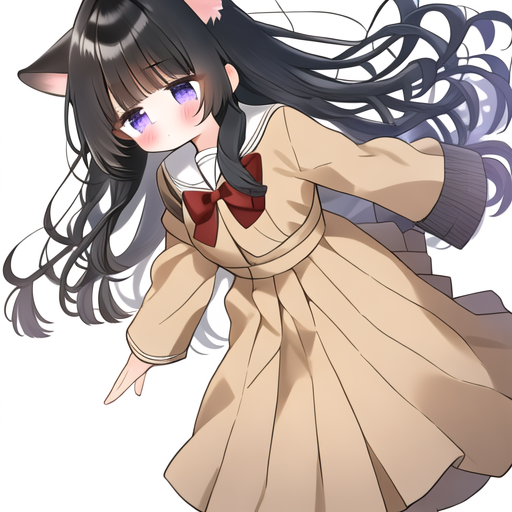}{\failimgw} &
    \FailMaskImg{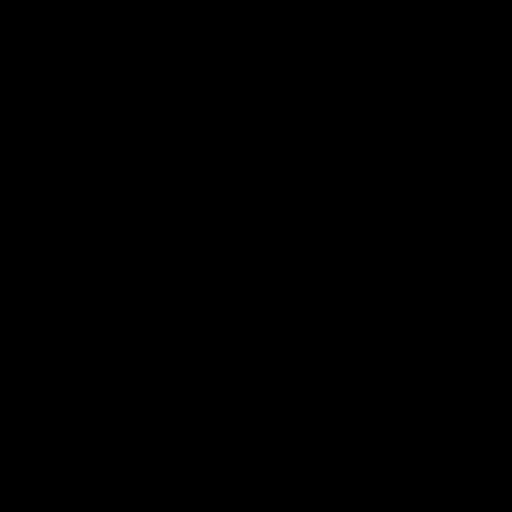}{\failimgw} \\
    (a1) & (a2) & (b1) & (b2) \\[0.18em]
    \multicolumn{4}{c}{%
      \begin{tabular}{@{}ccc@{}}
        \FailMaskImg{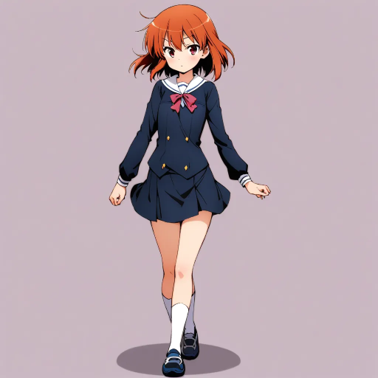}{\failimgw} &
        \FailMaskImg{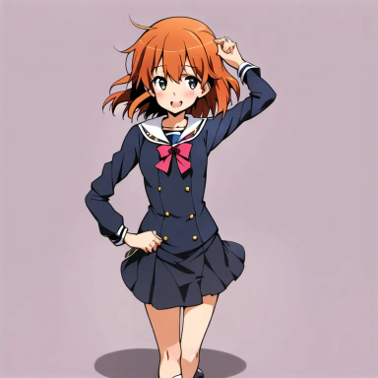}{\failimgw} &
        \FailMaskImg{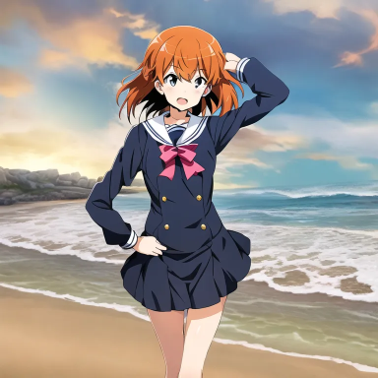}{\failimgw} \\
        (c1) & (c2) & (c3) \\
      \end{tabular}%
    } \\
  \end{tabular}
  \caption{Failed OpenPose-conditioning cases and SAM~3-based masking robustness.}
  \label{fig:supp_fail_mask}
\end{figure}

\cref{fig:supp_fail_mask} shows representative failure cases of AnimeAdapter under OpenPose guidance, together with an analysis of the robustness of our SAM~3-based masking.
The \emph{first row} highlights OpenPose-conditioning failures: panels (a1--a2) show generated multi-person artifacts even when the input prompt and OpenPose condition explicitly request a single subject; (b1--b2) show samples on which the pose detector fails to recover a skeleton from the generated image.
The \emph{second row} uses a single reference image to compare SAM~3 foreground masking (background suppressed) against an unmasked baseline: (c1) reference, (c2) generation without masking, and (c3) generation with masking.
SAM~3 already yields reliable segmentation; even in extreme cases that leave minor background leakage, our model remains visually stable.

\section{Data Safety Filtering and IP Concern}
\label{sec:supp_data_safety}
The $\sim$4.5M entries used to build our dataset are obtained from $\sim$9.1M raw Danbooru2025~\cite{danbooru2025metadata} records by removing NSFW content (\texttt{q}/\texttt{e} ratings) and non-single-subject posts, following common community filtering practice.
Our release policy is research-oriented and follows the dataset terms: we release only the permissible metadata and the processing pipeline, and do not redistribute restricted copyrighted originals.

{
    \small
    \bibliographystyle{ieeenat_fullname}
    \bibliography{main}
}